%% file: main.tex
\pdfoutput=1

\documentclass{article}

\usepackage{microtype}
\usepackage{graphicx}
\usepackage{subfigure}
\usepackage{array}
\usepackage{booktabs}
\usepackage{amsmath}
\usepackage{amssymb}
\usepackage{xcolor}
\usepackage{algorithm}
\usepackage{algorithmic}

\usepackage{hyperref}
\usepackage{xurl}

\usepackage[accepted]{mlsys2025}
\hypersetup{pdfsubject={},pdfkeywords={}}
\makeatletter
\renewcommand{\Notice@String}{Preprint.}
\renewcommand{\printAffiliationsAndNotice}[1]{%
\stepcounter{@affiliationcounter}%
{\let\thefootnote\relax\footnotetext{\hspace*{-\footnotesep}\ifdefined\isaccepted #1\fi%
\forloop{@affilnum}{1}{\value{@affilnum} < \value{@affiliationcounter}}{
\textsuperscript{\arabic{@affilnum}}\ifcsname @affilname\the@affilnum\endcsname%
\csname @affilname\the@affilnum\endcsname%
\else
{\bf AUTHORERR: Missing \textbackslash{}mlsysaffiliation.}
\fi
}.

\ \\
\Notice@String
}
}
}
\makeatother
\usepackage[capitalise]{cleveref}

\newcommand{\resBest}[1]{\textbf{#1}}
\newcommand{\resMatched}[1]{\underline{#1}}
\DeclareRobustCommand{\modelsep}{\tikz[baseline]\draw[black!40, line width=.4pt, dash pattern=on 1.5pt off 1.5pt] (0,-\dp\strutbox) -- (0,\ht\strutbox);}
\input{results/results_macros}
\input{results/quality_macros}
\input{results/calib_time_macros}

\newcommand{\methodname}{NestMSQ}

\usepackage{placeins}

\usepackage{tikz}
\usetikzlibrary{arrows.meta,positioning,fit}

\mlsystitlerunning{NestMSQ: Nested Multi-State Quantization for Fast and Accurate LLM Prefill and Decoding}

\begin{document}

\twocolumn[
\mlsystitle{NestMSQ: Nested Multi-State Quantization for Fast and Accurate LLM Prefill and Decoding}

\begin{mlsysauthorlist}
\mlsysauthor{Simon Kurz}{tud}
\mlsysauthor{Sebastian Buschj\"ager}{lamarr}
\mlsysauthor{Basile Lewandowski}{unine}
\mlsysauthor{Lydia Y. Chen}{unine,tudelft}
\mlsysauthor{Jian-Jia Chen}{rwth}
\end{mlsysauthorlist}

\mlsysaffiliation{tud}{TU Dortmund University, Dortmund, Germany}
\mlsysaffiliation{lamarr}{Lamarr Institute for Machine Learning and Artificial Intelligence, TU Dortmund University, Dortmund, Germany}
\mlsysaffiliation{unine}{University of Neuch\^atel, Neuch\^atel, Switzerland}
\mlsysaffiliation{tudelft}{Delft University of Technology, Delft, The Netherlands}
\mlsysaffiliation{rwth}{RWTH Aachen University, Aachen, Germany}

\mlsyskeywords{Post-training quantization, multi-precision inference, LLM serving}

\vskip 0.3in

\begin{abstract}

\input{sections/00_abstract}
\end{abstract}
]

\printAffiliationsAndNotice{}

\input{sections/01_introduction}
\input{sections/02_related_work}
\input{sections/03_background}
\input{sections/04_method}

\input{sections/05_experiments}

\input{sections/06_systems}
\input{sections/07_discussion}
\input{sections/08_conclusion}

\section*{Acknowledgements}
This work has received funding from the German Federal Ministry of Research, Technology and Space (BMFTR) in the course of the 6GEM+ transfer hub under grant number 16KIS2412. It has also been funded by the BMFTR and the state of North Rhine-Westphalia as part of the Lamarr Institute for Machine Learning and Artificial Intelligence.

\bibliography{references}
\bibliographystyle{mlsys2025}

\end{document}

%% file: results/results_macros.tex
\newcommand{\resLlamaNestMSQWFourASixteenAvg}{75.04}

%% file: results/quality_macros.tex
\newcommand{\resLlamaRtnWFourAFourCsr}{66.26}
\newcommand{\resLlamaRtnWFourAFourGsm}{55.80}
\newcommand{\resLlamaRtnWFourAFourMmlu}{53.94}
\newcommand{\resLlamaRtnWFourAFourIfeval}{68.27}
\newcommand{\resLlamaRtnWFourAFourAvg}{61.07}
\newcommand{\resLlamaRtnWEightAEightCsr}{73.57}
\newcommand{\resLlamaRtnWEightAEightGsm}{84.46}
\newcommand{\resLlamaRtnWEightAEightMmlu}{69.04}
\newcommand{\resLlamaRtnWEightAEightIfeval}{78.31}
\newcommand{\resLlamaRtnWEightAEightAvg}{76.35}
\newcommand{\resLlamaRtnWFourASixteenCsr}{71.98}
\newcommand{\resLlamaRtnWFourASixteenGsm}{79.15}
\newcommand{\resLlamaRtnWFourASixteenMmlu}{64.97}
\newcommand{\resLlamaRtnWFourASixteenIfeval}{79.72}
\newcommand{\resLlamaRtnWFourASixteenAvg}{73.96}
\newcommand{\resLlamaRtnWEightASixteenCsr}{73.76}
\newcommand{\resLlamaRtnWEightASixteenGsm}{85.44}
\newcommand{\resLlamaRtnWEightASixteenMmlu}{69.33}
\newcommand{\resLlamaRtnWEightASixteenIfeval}{78.31}
\newcommand{\resLlamaRtnWEightASixteenAvg}{76.71}
\newcommand{\resLlamaDedWFourAFourCsr}{70.00}
\newcommand{\resLlamaDedWFourAFourGsm}{74.22}
\newcommand{\resLlamaDedWFourAFourMmlu}{62.20}
\newcommand{\resLlamaDedWFourAFourIfeval}{73.27}
\newcommand{\resLlamaDedWFourAFourAvg}{69.92}
\newcommand{\resLlamaDedWEightAEightCsr}{73.72}
\newcommand{\resLlamaDedWEightAEightGsm}{85.60}
\newcommand{\resLlamaDedWEightAEightMmlu}{69.21}
\newcommand{\resLlamaDedWEightAEightIfeval}{78.32}
\newcommand{\resLlamaDedWEightAEightAvg}{76.71}
\newcommand{\resLlamaDedWFourASixteenCsr}{72.59}
\newcommand{\resLlamaDedWFourASixteenGsm}{82.34}
\newcommand{\resLlamaDedWFourASixteenMmlu}{66.98}
\newcommand{\resLlamaDedWFourASixteenIfeval}{76.70}
\newcommand{\resLlamaDedWFourASixteenAvg}{74.65}
\newcommand{\resLlamaDedWEightASixteenCsr}{73.60}
\newcommand{\resLlamaDedWEightASixteenGsm}{85.75}
\newcommand{\resLlamaDedWEightASixteenMmlu}{69.31}
\newcommand{\resLlamaDedWEightASixteenIfeval}{76.82}
\newcommand{\resLlamaDedWEightASixteenAvg}{76.37}
\newcommand{\resLlamaMatGptqWFourASixteenCsr}{72.57}
\newcommand{\resLlamaMatGptqWFourASixteenGsm}{81.96}
\newcommand{\resLlamaMatGptqWFourASixteenMmlu}{67.51}
\newcommand{\resLlamaMatGptqWFourASixteenIfeval}{76.34}
\newcommand{\resLlamaMatGptqWFourASixteenAvg}{74.60}
\newcommand{\resLlamaMatGptqWEightASixteenCsr}{73.64}
\newcommand{\resLlamaMatGptqWEightASixteenGsm}{84.91}
\newcommand{\resLlamaMatGptqWEightASixteenMmlu}{69.41}
\newcommand{\resLlamaMatGptqWEightASixteenIfeval}{78.86}
\newcommand{\resLlamaMatGptqWEightASixteenAvg}{76.71}
\newcommand{\resLlamaAnyPrecWFourASixteenCsr}{72.74}
\newcommand{\resLlamaAnyPrecWFourASixteenGsm}{80.44}
\newcommand{\resLlamaAnyPrecWFourASixteenMmlu}{67.40}
\newcommand{\resLlamaAnyPrecWFourASixteenIfeval}{77.38}
\newcommand{\resLlamaAnyPrecWFourASixteenAvg}{74.49}
\newcommand{\resLlamaAnyPrecWEightASixteenCsr}{73.80}
\newcommand{\resLlamaAnyPrecWEightASixteenGsm}{85.97}
\newcommand{\resLlamaAnyPrecWEightASixteenMmlu}{69.40}
\newcommand{\resLlamaAnyPrecWEightASixteenIfeval}{77.19}
\newcommand{\resLlamaAnyPrecWEightASixteenAvg}{76.59}
\newcommand{\resQwenRtnWFourAFourCsr}{63.20}
\newcommand{\resQwenRtnWFourAFourGsm}{63.46}
\newcommand{\resQwenRtnWFourAFourMmlu}{63.41}
\newcommand{\resQwenRtnWFourAFourIfeval}{77.52}
\newcommand{\resQwenRtnWFourAFourAvg}{66.90}
\newcommand{\resQwenRtnWEightAEightCsr}{71.66}
\newcommand{\resQwenRtnWEightAEightGsm}{87.64}
\newcommand{\resQwenRtnWEightAEightMmlu}{74.79}
\newcommand{\resQwenRtnWEightAEightIfeval}{85.68}
\newcommand{\resQwenRtnWEightAEightAvg}{79.94}
\newcommand{\resQwenRtnWFourASixteenCsr}{68.82}
\newcommand{\resQwenRtnWFourASixteenGsm}{84.76}
\newcommand{\resQwenRtnWFourASixteenMmlu}{72.08}
\newcommand{\resQwenRtnWFourASixteenIfeval}{83.67}
\newcommand{\resQwenRtnWFourASixteenAvg}{77.33}
\newcommand{\resQwenRtnWEightASixteenCsr}{71.62}
\newcommand{\resQwenRtnWEightASixteenGsm}{87.72}
\newcommand{\resQwenRtnWEightASixteenMmlu}{74.97}
\newcommand{\resQwenRtnWEightASixteenIfeval}{85.19}
\newcommand{\resQwenRtnWEightASixteenAvg}{79.88}
\newcommand{\resQwenDedWFourAFourCsr}{67.34}
\newcommand{\resQwenDedWFourAFourGsm}{81.05}
\newcommand{\resQwenDedWFourAFourMmlu}{69.31}
\newcommand{\resQwenDedWFourAFourIfeval}{81.70}
\newcommand{\resQwenDedWFourAFourAvg}{74.85}
\newcommand{\resQwenDedWEightAEightCsr}{71.53}
\newcommand{\resQwenDedWEightAEightGsm}{88.17}
\newcommand{\resQwenDedWEightAEightMmlu}{74.87}
\newcommand{\resQwenDedWEightAEightIfeval}{84.80}
\newcommand{\resQwenDedWEightAEightAvg}{79.84}
\newcommand{\resQwenDedWFourASixteenCsr}{70.21}
\newcommand{\resQwenDedWFourASixteenGsm}{85.82}
\newcommand{\resQwenDedWFourASixteenMmlu}{73.61}
\newcommand{\resQwenDedWFourASixteenIfeval}{86.42}
\newcommand{\resQwenDedWFourASixteenAvg}{79.02}
\newcommand{\resQwenDedWEightASixteenCsr}{71.54}
\newcommand{\resQwenDedWEightASixteenGsm}{87.49}
\newcommand{\resQwenDedWEightASixteenMmlu}{74.92}
\newcommand{\resQwenDedWEightASixteenIfeval}{84.98}
\newcommand{\resQwenDedWEightASixteenAvg}{79.73}
\newcommand{\resQwenMatGptqWFourASixteenCsr}{71.42}
\newcommand{\resQwenMatGptqWFourASixteenGsm}{86.88}
\newcommand{\resQwenMatGptqWFourASixteenMmlu}{73.98}
\newcommand{\resQwenMatGptqWFourASixteenIfeval}{83.49}
\newcommand{\resQwenMatGptqWFourASixteenAvg}{78.94}
\newcommand{\resQwenMatGptqWEightASixteenCsr}{71.45}
\newcommand{\resQwenMatGptqWEightASixteenGsm}{87.34}
\newcommand{\resQwenMatGptqWEightASixteenMmlu}{75.16}
\newcommand{\resQwenMatGptqWEightASixteenIfeval}{85.04}
\newcommand{\resQwenMatGptqWEightASixteenAvg}{79.75}
\newcommand{\resQwenAnyPrecWFourASixteenCsr}{71.16}
\newcommand{\resQwenAnyPrecWFourASixteenGsm}{86.28}
\newcommand{\resQwenAnyPrecWFourASixteenMmlu}{73.93}
\newcommand{\resQwenAnyPrecWFourASixteenIfeval}{82.82}
\newcommand{\resQwenAnyPrecWFourASixteenAvg}{78.55}
\newcommand{\resQwenAnyPrecWEightASixteenCsr}{71.70}
\newcommand{\resQwenAnyPrecWEightASixteenGsm}{87.04}
\newcommand{\resQwenAnyPrecWEightASixteenMmlu}{74.92}
\newcommand{\resQwenAnyPrecWEightASixteenIfeval}{84.01}
\newcommand{\resQwenAnyPrecWEightASixteenAvg}{79.42}

%% file: results/calib_time_macros.tex
\newcommand{\resCalibTimeMsGptaq}{3\,h\,8\,min}
\newcommand{\resCalibTimeSingleFlow}{2\,h\,34\,min}

%% file: sections/00_abstract.tex
Nested quantization of large language models embeds lower-precision models in the bits of a higher-precision one, so one checkpoint serves requests with different latency and accuracy needs without deploying a separate model for each precision. We present \methodname{} (Nested Multi-State Quantization), a post-training quantization (PTQ) method whose INT8 checkpoint provides 4-bit weights for speed and 8-bit weights for accuracy, each with low-bit activations for arithmetic-heavy long prefill (W4A4, W8A8) or with 16-bit activations for bandwidth-heavy low-batch decoding (W4A16, W8A16). The shared checkpoint can switch precision per request and between prefill and decoding without loading another model. During calibration, different weight and activation precisions give the four modes different layer inputs and rounding errors. Compensation-based PTQ offsets rounding errors by adjusting weights still to be quantized. Nested weight precisions already favor different adjustments, while varying activation precision also makes the modes evaluate these errors on different inputs. \methodname{} therefore calibrates each mode on its own inputs and keeps its preferred adjustments separate while choosing each INT8 weight jointly for all modes. Compared with sharing inputs and adjustments, this closes 73\,\% of the W4A4 perplexity gap to a separately quantized model. Across two 8B models, the nested checkpoint stays within 1.3 average-accuracy points of separately quantized models in every mode. For single requests, W4A16 decoding after a W4A4 prefill keeps W4A4 speed and matches W4A16 accuracy on MATH-500. When long-input W4A4 prefill costs accuracy, W8A8 prefill with W4A16 decoding stays within 2.6 points of BF16 on three tasks at 1.96 to 2.61 times its output throughput. This flexibility extends to continuous-batching serving, where precision adapts as the number of parallel requests changes.

%% file: sections/01_introduction.tex
\section{Introduction}
\label{sec:introduction}

Quantization makes it possible to run large language models (LLMs) on a single GPU with limited memory, such as a 24\,GB consumer GPU~\cite{frantar2023optq,lin2024awq}. The precision chosen at quantization time, however, sets one speed and one accuracy for every request, whereas applications and requests differ in the speed and accuracy they need~\cite{park2024anyprecisionllm,kleinegger2026matgptqv2,park2026anybcq,kim2026dpamoe}. An interactive reply may favor speed, while a math or coding task may favor accuracy. A second model for the other need occupies memory that the KV cache would otherwise use~\cite{vllm}, which limits how many requests the GPU serves at once or how long their sequences can be. Nested quantization avoids the need for a second model. It embeds lower-precision models in the bits of a higher-precision one~\cite{nair2025matryoshka,park2024anyprecisionllm,kleinegger2026matgptqv2}, so one checkpoint provides several precision choices at the memory cost of the largest model. We introduce \methodname{} (Nested Multi-State Quantization), a post-training quantization (PTQ) method for such nested checkpoints.

Which precision is fast depends on the phase of inference. Decoding few requests is bound by loading weights, so 4-bit weights with 16-bit activations suffice~\cite{pope2023scaling}, whereas long prefill is bound by arithmetic and is fast only with low-bit activations on integer Tensor Cores~\cite{nvidia2023ada,ashkboos2024quarot,lin2025qserve}. One request therefore needs different precisions over its lifetime, and the best precision for decoding changes again with load, so a single fixed precision gives up speed or accuracy for part of the work. Many nested methods quantize only the weights~\cite{nair2025matryoshka,park2024anyprecisionllm,kleinegger2026matgptqv2}, so their fast model barely speeds up prefill. Some methods serve several precisions with quantized activations but give up either low-bit prefill arithmetic or the memory savings of a fully quantized checkpoint. BitNest~\cite{yang2026bitnest} prefills with 16-bit arithmetic, whereas NestedFP~\cite{lee2026nestedfp} and QuEPT~\cite{xu2026quept} retain a high-precision weight representation. \methodname{} instead stores 8-bit weights that also serve as 4-bit weights and runs both with 16-bit activations or with activations of the same bit width, which gives the modes W4A16, W8A16, W4A4 and W8A8. The two WxAx modes match established weight-activation settings and integer Tensor Core formats~\cite{xiao2023smoothquant,ashkboos2024quarot,nvidia2023ada}.

Fast W4A4 prefill requires PTQ that handles quantized activations, which GPTAQ~\cite{li2025gptaq} provides but with one model per precision. A single nested checkpoint additionally requires W4 and W8 to be calibrated jointly, which MatGPTQ~\cite{kleinegger2026matgptqv2} provides for weight-only modes. MatGPTQ compensates each rounding error by adjusting weights still to be quantized, while all precisions share the same layer inputs and remaining weights. W4 and W8 already induce different rounding errors and therefore prefer different adjustments. Varying activation precision also gives each mode different layer inputs, so the modes evaluate these errors differently. One shared set of remaining weights can only average these mode-specific adjustments (Fig.~\ref{fig:compensation}).

\methodname{} therefore calibrates each mode on its own layer inputs and keeps a separate copy of the remaining weights for each mode. We call these the calibration flow and the working state of the mode. Each mode can thus apply its own adjustment instead of the average, while every stored INT8 weight is chosen jointly for all modes. This separation exists only during PTQ. At inference, all four modes use the same checkpoint, which makes runtime switching possible without another model.

Our contributions are as follows.
\begin{itemize}
\item \textbf{\methodname{}.}
A nested PTQ method that calibrates one checkpoint of 8-bit weights for W4A4, W8A8, W4A16 and W8A16 with a calibration flow and a working state per mode. Together, the per-mode flows and working states close 73\,\% of the W4A4 perplexity gap from one shared flow and state to a separately quantized model (Table~\ref{tab:ladder_ppl}). In average accuracy, the checkpoint stays within 0.7 points of dedicated GPTQ and GPTAQ models in three modes on Llama-3.1-8B-Instruct and Qwen3-8B, and within 1.3 points at W4A4 (Table~\ref{tab:main}).

\item \textbf{Flexibility from one checkpoint.}
Precision can be raised where it improves accuracy and lowered where it buys speed, both across prefill and decoding and as serving load changes. W4A16 decoding after W4A4 prefill keeps W4A4 speed and matches W4A16 accuracy on MATH-500. When prompt processing needs higher precision, W8A8 prefill with W4A16 decoding stays within 2.6 points of BF16 across the three tasks of Table~\ref{tab:switch} at 1.96 to 2.61 times its speed, which no fixed mode reaches. Under continuous batching in our vLLM integration, the same checkpoint adapts its activation precision to the load and is more accurate than fixed W4A4 on MATH-500 and GSM8K at 0.96 to 1.05 times its throughput (Fig.~\ref{fig:serving_load}, Table~\ref{tab:load_switch}).
\end{itemize}

%% file: sections/02_related_work.tex
\section{Related Work}
\label{sec:related_work}

Dedicated quantization methods optimize one model for one precision setting, and they fall into the two regimes our checkpoint has to serve. Weight-only methods (WxA16) reduce the weight traffic that dominates memory-bound decode. They store integer codes~\cite{frantar2023optq,lin2024awq} or indices into non-uniform or lattice codebooks~\cite{kim2024squeezellm,tseng2024quipsharp}, whose kernels target matrix--vector products. Weight-activation methods (WxAx) also quantize the activations, so that compute-bound matrix multiplications run in low-precision arithmetic. At 8 bits, methods commonly handle activation outliers~\cite{dettmers2022llmint8,xiao2023smoothquant}, and at 4 bits they often rotate weights and activations~\cite{ashkboos2024quarot,liu2025spinquant}. Each of these methods commits to one regime when the model is compressed. A WxA16 model gains little speed in compute-bound prefill, whereas a WxAx model keeps its low-bit activations in decode, where they bring little speed but can cost accuracy. Serving both phases well thus takes one model per regime. Most of them are post-training methods that need only a small calibration set, whereas dedicated quantization-aware training~\cite{liu2024llmqat,chen2025efficientqat} retrains the model for every precision. GPTQ~\cite{frantar2023optq} and GPTAQ~\cite{li2025gptaq} serve as our dedicated references for the WxA16 and the WxAx modes.

Nested and any-precision quantization keeps several weight precisions in one stored model. Training-based methods learn such a model directly. Any-Precision DNNs~\cite{yu2021anyprecisiondnns} train networks that run at several bit-widths, and MatQuant~\cite{nair2025matryoshka} co-trains an INT8 model whose most significant bits form INT4 and INT2 views, on top of QAT or OmniQuant~\cite{shao2024omniquant}. Its lower views round this 8-bit code rather than truncate it~\cite{nair2025matryoshka}, so an $r$-bit view also reads the next bit and cannot be served from its own $r$ bits~\cite{kleinegger2026matgptqv2}. MF-QAT~\cite{xu2026mfqat} trains one 8-bit microscaling model and derives lower formats with a conversion step instead of a bit prefix.
Post-training methods avoid this cost. Any-Precision LLM~\cite{park2024anyprecisionllm} upscales a low-bit post-training model into overlaid 3- to 8-bit variants, MatGPTQ~\cite{kleinegger2026matgptqv2} adapts GPTQ~\cite{frantar2023optq} to optimize one stored 8-bit model jointly for all precisions read as exact bit prefixes of it, and GRINQH~\cite{oberlander2026grinqh} truncates such a stored model in 2-bit steps. RRQ~\cite{luo2026rrq} and StagQ~\cite{wei2026stagq} extend a 2-bit base by stages that quantize the remaining error.
Any-Precision LLM, AnyBCQ~\cite{park2026anybcq} and MoBiQuant~\cite{wang2026mobiquant} build their views from bit-planes or residual slices. The kernels of Any-Precision LLM and AnyBCQ serve only small-batch decode and fall back to dequantization and a 16-bit matrix multiplication in prefill~\cite{kleinegger2026matgptqv2}.
All of these methods vary only the weight precision and keep 16-bit activations, so none of them uses low-bit arithmetic when compute dominates.

Methods that serve several precisions with low-bit activations differ in which combinations they cover and in how they obtain them. BitNest~\cite{yang2026bitnest} and, for vision models, NestQuant~\cite{xie2025nestquant} run both of their weight precisions with one fixed activation precision, QSpec~\cite{zhao2025qspec} runs its 4-bit weights with both 4-bit and 16-bit activations, and QuEPT~\cite{xu2026quept} evaluates coupled weight and activation bit widths on LLMs, and NestedFP~\cite{lee2026nestedfp} lowers both together from FP16 to FP8. Disaggregated quantization~\cite{panferov2026disaggregated} specializes one low-precision operating point to each phase, using low-bit weight-activation compute for prefill and compact weight-only decoding, from either one stored weight representation or separate prefill weights. BitNest and QSpec use one stored model for speculative decoding, in which the lower-precision mode drafts tokens and the higher-precision mode checks them. BitNest rounds the remaining error of a 4-bit GPTQ model to four additional bits without calibrating the 8-bit result, and its current implementation prefills with 16-bit arithmetic. NestedFP nests FP8 in FP16 weights and therefore retains a 16-bit checkpoint. DPS~\cite{nguyen2026dps} instead switches between nested FP16 and FP8 weights under KV-cache pressure and repurposes the residual-weight memory for KV-cache blocks. QuEPT calibrates several bit widths jointly on shared full-precision weights, so its modes do not come from one shared quantized code, and it reports accuracy without low-bit kernels. During calibration, it merges the outputs of several bit widths into one input for the next block.
MatGPTQ calibrates all its precisions on one shared flow, so neither propagates the calibration data separately through the quantized network of each precision setting.

Runtime precision switching chooses among the precisions a model keeps resident. DP-LLM~\cite{kwon2025dpllm} assigns a precision to each layer based on the input, MoBiQuant~\cite{wang2026mobiquant} selects a precision per token and GRINQH~\cite{oberlander2026grinqh} per input channel, and PMPD~\cite{chen2025pmpd} lowers the precision as decoding proceeds. FlexQuant~\cite{liu2025flexquant} prefills in W8A8 and replaces the INT8 weights of individual layers by separately stored INT4 weights during decoding.
These methods switch among weight precisions only. We additionally switch the activation precision with the load of each iteration, using low-bit activations when arithmetic limits speed and 16-bit activations when weight loading limits it.

%% file: sections/03_background.tex
\section{Background}
\label{sec:background}

\subsection{Post-Training Quantization and Sequential Compensation}
\label{subsec:background_gptq}

Our method builds on GPTQ~\cite{frantar2023optq}, which alternates quantization and compensation, so every quantization decision changes the weights from which later decisions are made. GPTQ quantizes a trained model one linear layer at a time with a small calibration set and minimizes the error of each layer's output on the observed inputs. Each output feature of the layer depends only on one row of the weight matrix, so the loss is a sum of independent per-row terms, and we write it for one row. For a weight row $\mathbf w\in\mathbb R^{1\times K}$ over $K$ input features, its quantized counterpart $\widehat{\mathbf w}\in\mathcal G^{1\times K}$ on a quantization grid $\mathcal G$, and the layer's inputs $\mathbf X\in\mathbb R^{K\times n}$ on $n$ calibration tokens, the row-wise loss $L$ and its minimization are

\begin{equation}
\begin{aligned}
\min_{\widehat{\mathbf w}}\;L(\widehat{\mathbf w})
&=
\min_{\widehat{\mathbf w}}\;
\bigl\|\widehat{\mathbf w}\mathbf X-\mathbf w\mathbf X\bigr\|_F^2\\
&=
\min_{\widehat{\mathbf w}}\;
\frac12(\widehat{\mathbf w}-\mathbf w)\,\mathbf H\,(\widehat{\mathbf w}-\mathbf w)^\top.
\end{aligned}
\label{eq:gptq_objective}
\end{equation}

The first form states what is reconstructed, namely the layer output on the calibration inputs. The second exposes $\mathbf H=2\mathbf X\mathbf X^\top$, the Hessian of this loss with respect to the weights. The inputs $\mathbf X$ are obtained by propagating the calibration samples through the already quantized preceding layers, so each layer is calibrated on the inputs the quantized network actually produces. We call this sequence of layer inputs the \emph{calibration flow}. $\mathbf H$ is computed from these inputs in full precision before the layer is quantized and is not part of the checkpoint.

GPTQ solves Eq.~\eqref{eq:gptq_objective} greedily and quantizes the row one input index $j=0,\ldots,K-1$ at a time. At step $j$, let $\mathcal F=\{k\in\{0,\ldots,K-1\}:k>j\}$ be the indices that are still unquantized, and $\mathbf H^{-1}$ the inverse of the submatrix of $\mathbf H$ on the rows and columns $\{j\}\cup \mathcal F$. It changes with $j$, which we leave implicit in the notation. GPTQ fixes $\widehat w_j$ at the value in $\mathcal G$ with the smallest loss increase and then compensates its error $e_j=w_j-\widehat w_j$ in the remaining weights,

\begin{equation}
\begin{aligned}
\widehat w_j
&=\operatorname*{arg\,min}_{g\in\mathcal G}
\frac{(g-w_j)^2}{[\mathbf H^{-1}]_{jj}},\\
\Delta\mathbf w_{\mathcal F}
&=-\frac{e_j}{[\mathbf H^{-1}]_{jj}}\,\mathbf H^{-1}_{j\mathcal F},
\qquad
\mathbf w_{\mathcal F}\leftarrow\mathbf w_{\mathcal F}+\Delta\mathbf w_{\mathcal F},
\end{aligned}
\label{eq:gptq_step}
\end{equation}

where $\mathbf H^{-1}_{j\mathcal F}$ is row $j$ of $\mathbf H^{-1}$ restricted to $\mathcal F$. GPTQ obtains all required rows of $\mathbf H^{-1}$ from one Cholesky factorization, which makes the procedure tractable at LLM scale.

The update turns $\mathbf w_{\mathcal F}$ into a \emph{working state}. It is neither the original weights nor yet the quantized ones, and every later decision is made from it. The working state carries the history of all earlier compensations. It exists only during quantization, and the checkpoint stores only the codes it leads to.

\subsection{Nested Multi-Precision GPTQ}
\label{subsec:background_matgptq}

MatGPTQ~\cite{kleinegger2026matgptqv2} extends GPTQ to several weight precisions read from one stored parent code, which is the nesting we build on. Let $\mathcal Q_c=\{0,\ldots,2^c-1\}$ be the set of $c$-bit parent codes and $\mathcal R$ the target weight precisions. The $r$-bit view of a parent code $q$ keeps its $r$ most significant bits,

\begin{equation}
\begin{aligned}
S'(q,r)&=\left\lfloor\frac{q}{2^{c-r}}\right\rfloor2^{c-r},
\qquad
\delta_r=\frac{2^{c-r}-1}{2},\\
\widehat w_r(q)&=s\bigl(S'(q,r)+\delta_r-z\bigr),
\end{aligned}
\label{eq:matgptq_slice}
\end{equation}

with a scale $s\in\mathbb R^{+}$ and a zero point $z\in\mathbb R$ shared by all views. Every lower-precision view is a prefix of the same stored code, so the precision can be selected at runtime without requantizing the model or storing separate weights. Truncation maps all codes that share a prefix to the smallest of them, and the offset $\delta_r$ moves the reconstruction to their midpoint and removes this systematic downward bias.

MatGPTQ keeps one working state for all views, which we write $\mathbf w^{\mathrm{sh}}$. Because every view reads the same code, MatGPTQ chooses the code at index $j$ jointly for all target precisions from the working weight $w_j^{\mathrm{sh}}$. It weights each precision $r$ with a factor $\lambda_r>0$. With equal factors, as in the code selection of \methodname{}, the selection is

\begin{equation}
q_j^\star
=\operatorname*{arg\,min}_{q\in\mathcal Q_c}
\sum_{r\in\mathcal R}
\bigl(\widehat w_r(q)-w_j^{\mathrm{sh}}\bigr)^2 .
\label{eq:matgptq_code}
\end{equation}

In its released code, MatGPTQ propagates the calibration data through its highest-precision view. All views thus share one calibration flow and one Hessian, and the GPTQ factor $1/[\mathbf H^{-1}]_{jj}$ of Eq.~\eqref{eq:gptq_step} is common to all terms and drops out.

The committed code leaves a different residual $e_{j,r}=w_j^{\mathrm{sh}}-\widehat w_r(q_j^\star)$ in every view, and each view would compensate its own residual with its own GPTQ update. Because MatGPTQ keeps only one working state, it applies the mean $\Delta\mathbf w_{\mathcal F}^{\mathrm{sh}}$ of these preferred updates, weighted with the same factors $\lambda_r$, to $\mathbf w^{\mathrm{sh}}$, which for equal factors is

\begin{equation}
\Delta\mathbf w_{\mathcal F,r}^{\star}
=-\frac{e_{j,r}}{[\mathbf H^{-1}]_{jj}}\,\mathbf H^{-1}_{j\mathcal F},
\quad
\Delta\mathbf w_{\mathcal F}^{\mathrm{sh}}
=\frac{1}{|\mathcal R|}\sum_{r\in\mathcal R}\Delta\mathbf w_{\mathcal F,r}^{\star}.
\label{eq:matgptq_shared_update}
\end{equation}

MatGPTQ thus shares three objects across its views, the stored parent code, the calibration flow and the temporary working state.

%% file: sections/04_method.tex
\section{\methodname{}}
\label{sec:method}

\methodname{} calibrates one nested 8-bit parent code for the four modes W4A4, W8A8, W4A16 and W8A16. The modes differ in weight and in activation precision, and both differences create conflicts that the calibration must resolve. The activation precision gives every mode its own layer inputs, which \methodname{} captures in one calibration flow per mode (Sec.~\ref{subsec:four_flows}). Together with the two weight precisions, these flows make the modes prefer different updates of the not yet quantized weights at every greedy PTQ step (Sec.~\ref{subsec:conflicting}). Section~\ref{subsec:mode_consistent} resolves this second conflict via per-mode working weights.

\subsection{Two Precision Axes, One Calibration Problem}
\label{subsec:four_flows}

The four modes add an activation axis to the weight axis of nested weight-only quantization and pair both 16-bit and low-bit activations with a fast 4-bit and an accurate 8-bit weight precision. We build on MatGPTQ to obtain the 4-bit and 8-bit weights as views (Eq.~\eqref{eq:matgptq_slice}) of one 8-bit parent code, read at $r\in\mathcal R=\{4,8\}$ of its 8 bits. So that the WxAx modes can feed both views directly to integer Tensor Cores, we store the parent code as a signed INT8 value $q=16C+R$, whose upper four bits form the signed INT4 value $C\in\{-8,\ldots,7\}$ and whose lower four bits form the refinement $R\in\{0,\ldots,15\}$. With one scale $s_8$ per output channel, the two views are

\begin{equation}
\widehat w_8(q)=s_8\,q,
\qquad
\widehat w_4(q)=s_8\!\left(16C+\tfrac{15}{2}\right),
\label{eq:w4_w8}
\end{equation}

which is Eq.~\eqref{eq:matgptq_slice} for the unsigned code $q+128$ and the zero point $z=128$. From here on, $q$ denotes this signed code. W4 reads only $C$, and W8 reads all of $q$.

Supporting 4-bit activations requires a PTQ method designed for W4A4. We therefore build the calibration of all four modes on GPTAQ~\cite{li2025gptaq}. Like GPTAQ, all four modes multiply weights and activations by the same fixed randomized Hadamard rotation of QuaRot~\cite{ashkboos2024quarot}, which spreads activation outliers over many entries and makes 4-bit activations easier to quantize. The modes must share this rotation, because they read the same stored code. GPTAQ also changes the calibration target. GPTQ (Eq.~\eqref{eq:gptq_objective}) gives the quantized and the original weights the same input $\mathbf X$, which already carries the errors of the quantized preceding layers. GPTAQ instead targets the output $\mathbf w\widetilde{\mathbf X}$ of the full-precision model, where $\widetilde{\mathbf X}$ is the layer's input when the calibration samples pass through the unquantized model,

\begin{equation}
\min_{\widehat{\mathbf w}}\;
\bigl\|\widehat{\mathbf w}\mathbf X-\mathbf w\widetilde{\mathbf X}\bigr\|_F^2,
\label{eq:gptaq_single}
\end{equation}

so that each layer also compensates the error accumulated upstream. Because $\widetilde{\mathbf X}$ does not depend on the mode, the full-precision output also serves as the common reference for all four modes.

A single calibration flow cannot distinguish W4A4 from W4A16 or W8A8 from W8A16, because each pair reads the same weights on the same inputs. The activation precision would therefore disappear from the calibration problem. We instead propagate the calibration samples separately through the quantized network of each mode, which gives mode $m$ its own calibration flow $\mathbf X_m$ and Hessian $\mathbf H_m$. The flows differ from the first layer on, because each mode quantizes its own activations.

The checkpoint stores one row of parent codes $\mathbf q$ per row of the weight matrix, which mode $m$ reads at its weight precision $r_m$. With all modes weighted equally, the calibration problem is

\begin{equation}
\begin{aligned}
\mathbf q^{\star}
&=\operatorname*{arg\,min}_{\mathbf q}
\sum_{m\in\mathcal M}
\bigl\|\widehat{\mathbf w}_{r_m}(\mathbf q)\mathbf X_m-\mathbf w\widetilde{\mathbf X}\bigr\|_F^2,\\
\mathcal M&=\{\mathrm{W4A4},\mathrm{W8A8},\mathrm{W4A16},\mathrm{W8A16}\},\\
\mathbf H_m&=2\mathbf X_m\mathbf X_m^{\top}.
\end{aligned}
\label{eq:gptaq_objective}
\end{equation}

The four networks are coupled only through $\mathbf q$, which carries no mode index because it is deployed. Only $\mathbf X_m$ and $\mathbf H_m$ are specific to mode $m$. Like GPTQ and GPTAQ, we solve this problem greedily, one input index at a time.

\input{figures/compensation}

\subsection{Shared Code, Conflicting Corrections}
\label{subsec:conflicting}

A GPTQ step selects a code and then compensates its error (Eq.~\eqref{eq:gptq_step}). MatGPTQ runs the selection and the compensation on one flow, and the per-mode flows change both. Like MatGPTQ's views, the four modes share one working state $\mathbf w^{\mathrm{sh}}$ (Sec.~\ref{subsec:background_matgptq}), and the code at step $j$ is chosen jointly for all modes from its working weight $w_j^{\mathrm{sh}}$. For a candidate $q$, a mode with weight precision $r$ causes the residual $e_{j,r}(q)=w_j^{\mathrm{sh}}-\widehat w_r(q)$ of Sec.~\ref{subsec:background_matgptq}, and the multi-mode selection sums the GPTQ selection cost of Eq.~\eqref{eq:gptq_step} over the modes,

\begin{equation}
q_j^\star
=\operatorname*{arg\,min}_{q\in\{-128,\ldots,127\}}
\sum_{m\in\mathcal M}
\frac{e_{j,r_m}(q)^2}{[\mathbf H_m^{-1}]_{jj}}.
\label{eq:mode_curvature_code}
\end{equation}

Unlike in Eq.~\eqref{eq:matgptq_code}, the factors $1/[\mathbf H_m^{-1}]_{jj}$ differ between modes and no longer drop out. They come from the flow of each mode and thus carry its activation precision and the errors propagated from earlier layers. W4A4 and W4A16 thus cause the same residual $e_{j,4}$ but weight it differently.

After $q_j^\star$ is committed, each mode prefers its own GPTAQ update of the remaining weights,

\begin{equation}
\begin{aligned}
\Delta\mathbf w_{\mathcal F,m}
&=-\frac{e_{j,r_m}(q_j^\star)}{[\mathbf H_m^{-1}]_{jj}}\,\mathbf H^{-1}_{m,j\mathcal F}
+w_j^{\mathrm{sh}}\,\mathbf P_{m,j\mathcal F},\\
\mathbf P_{m,j\mathcal F}
&=\alpha\,\mathbf G_{m,j\mathcal F}\,\mathbf H_{m,\mathcal F\mathcal F}^{-1},\\
\mathbf G_m
&=2(\widetilde{\mathbf X}-\mathbf X_m)\mathbf X_m^\top.
\end{aligned}
\label{eq:gptaq_mode_update}
\end{equation}

The first term is the GPTQ compensation of the committed residual under the mode's own Hessian. The second term is GPTAQ's compensation of the input error that the mode's flow has inherited from earlier layers, scaled by $\alpha$.

The four modes thus prefer different updates, because W4 and W8 read different values from the same code and because each flow has its own Hessian and upstream error. MatGPTQ's single working state cannot follow these distinct preferences, because it can apply only one update (\textcircled{\scriptsize 1} in Fig.~\ref{fig:compensation}). The shared-state baseline applies their mean, as MatGPTQ does for its views (Eq.~\eqref{eq:matgptq_shared_update}). The mean averages away the preference of each mode, so the modes carry an additional loss into the next step. Even with MatGPTQ's single flow, it leaves the additional loss, averaged over the two views,

\begin{equation}
\Delta_{\mathrm{state}}
=\frac{\kappa_j}{8}\,\bigl(e_{j,4}-e_{j,8}\bigr)^2\;\ge\;0,
\label{eq:state_penalty_variance}
\end{equation}

where $\kappa_j\ge0$ depends only on $\mathbf H$. The residual difference never vanishes, because W4 discards the lower four bits $R$ that W8 keeps. For every code, $e_{j,4}-e_{j,8}=s_8(R-\tfrac{15}{2})\neq0$, so every step with $\kappa_j>0$ pays at least $\kappa_j s_8^2/32$.

\subsection{One Working State per Mode}
\label{subsec:mode_consistent}

Rather than searching for a better common update, \methodname{} removes the requirement for one. The parent code must be common because it is deployed. The working state is temporary, so nothing forces it to be common. \methodname{} therefore keeps one working state $\mathbf w^{(m)}$ per mode and applies each mode's update to its own state. Eqs.~\eqref{eq:mode_curvature_code} and~\eqref{eq:gptaq_mode_update} keep their form, but mode $m$ reads its own working weight $w_j^{(m)}$ in place of $w_j^{\mathrm{sh}}$. Its residual then depends on the mode and no longer only on its weight precision. We therefore write $e_{m,j}(q)=w_j^{(m)}-\widehat w_{r_m}(q)$ in place of $e_{j,r_m}(q)$ in both equations. The shared-state baseline of Sec.~\ref{subsec:conflicting} instead applies the mean of the four updates to $\mathbf w^{\mathrm{sh}}$,

\begin{equation}
\begin{aligned}
\text{baseline}\quad
\mathbf w_{\mathcal F}^{\mathrm{sh}}
&\leftarrow
\mathbf w_{\mathcal F}^{\mathrm{sh}}
+\frac{1}{|\mathcal M|}
\sum_{m\in\mathcal M}\Delta\mathbf w_{\mathcal F,m},\\
\text{\methodname{}}\quad
\mathbf w_{\mathcal F}^{(m)}
&\leftarrow
\mathbf w_{\mathcal F}^{(m)}+\Delta\mathbf w_{\mathcal F,m}
\quad\forall m\in\mathcal M.
\end{aligned}
\label{eq:working_state}
\end{equation}

Figure~\ref{fig:compensation} compares \methodname{} with the shared-state baseline on one projection. Both runs use the same per-mode flows and the same selection rule (Eq.~\eqref{eq:mode_curvature_code}) and differ only in the working state (\textcircled{\scriptsize 1}), so any difference between them comes from the working state alone. Averaging the preferred updates makes the shared working state drift from the per-mode states (\textcircled{\scriptsize 2}). Because later quantized weights are chosen from the current working weights, this drift changes future choices and raises the committed loss (\textcircled{\scriptsize 3}). We write the rows of the projection as a matrix $\mathbf W$, whose rows share the Hessian $\mathbf H_m$ of mode $m$. After step $j$, we measure the drift and the committed loss as

\begin{equation}
\begin{aligned}
D_m(j)&=\operatorname{RMS}\bigl(\mathbf W_{\mathcal F}^{(m)}-\mathbf W_{\mathcal F}^{\mathrm{sh}}\bigr),\\
L_{a,m}(j)&=\bigl\|\widehat{\mathbf W}_{\le j}^{a,m}\mathbf X_m-\mathbf W\widetilde{\mathbf X}\bigr\|_F^2.
\end{aligned}
\label{eq:fig_diagnostics}
\end{equation}

$D_m(j)$ compares the shared working state of the baseline with the working state of mode $m$ in \methodname{} on the remaining columns $\mathcal F$. Both come from separate calibration runs at the same step $j$. $L_{a,m}(j)$ is the loss of the layer output of mode $m$ against the full-precision model (Eq.~\eqref{eq:gptaq_objective}), with the columns up to $j$ quantized by \methodname{} ($a=\mathrm{MS}$) or by the baseline ($a=\mathrm{sh}$). The drift grows over the pass, most steeply in the last columns (\textcircled{\scriptsize 2}), and the committed-loss gap $L_{\mathrm{sh},m}(j)-L_{\mathrm{MS},m}(j)$ grows with it (\textcircled{\scriptsize 3}). The W4 modes separate more strongly than the W8 modes. Per-mode working states thus commit less loss in every mode. In all 140 cases of 35 measured projections and four modes, the per-mode working states end with the lower loss.

\begin{table}[t]
\caption{Contribution of per-mode flows and per-mode working states, alone and combined. WikiText-2 perplexity of Llama-3.1-8B-Instruct (lower is better). One dot means that the four modes share one calibration flow or working state, four dots that each mode keeps its own. Bold marks the best quantized method and underline the best one-checkpoint method. Per-mode flows with a shared state form the shared-state baseline.}
\label{tab:ladder_ppl}
\centering
\fontsize{8pt}{8.9pt}\selectfont \setlength{\tabcolsep}{2.6pt}\input{results/ladder_ppl_table}
\end{table}

Table~\ref{tab:ladder_ppl} measures on the whole model what the per-mode flows and working states contribute. Starting from MatGPTQ's structure, with one shared flow and one shared working state, we give each mode its own flow, its own working state or both. Per-mode flows alone lower the W4A4 perplexity but raise it at W8, whereas per-mode working states alone help in every mode. Both together close 73\,\% of the W4A4 gap to the dedicated GPTAQ model, so the per-mode flows pay off once each mode also keeps its own working state.

%% file: figures/compensation.tex
\begin{figure*}[t]
\centering
\begin{minipage}[c]{0.43\textwidth}
\centering
\begin{tikzpicture}[font=\fontsize{7.35}{8.2}\selectfont,
>={Stealth[length=3.5pt]},
num/.style={draw, circle, minimum size=8.5pt, inner sep=0pt,
font=\fontsize{7.35}{8.2}\selectfont\bfseries, fill=white}]
\draw[black!60, densely dashed, thick] (2.95,2.45) -- (2.95,1.83)
(6.6,2.45) -- (6.6,1.83);
\node[num] at (-0.05,3.17) {3};
\node[anchor=west, font=\fontsize{7.35}{8.2}\selectfont\itshape] at (0.12,3.17)
{columns of $W$ at step $j$};
\node[num] at (6.45,3.17) {2};
\fill[green!25] (0,2.45) rectangle (2.6,2.9);
\fill[orange!45] (2.6,2.45) rectangle (2.95,2.9);
\fill[gray!10] (2.95,2.45) rectangle (6.6,2.9);
\draw (0,2.45) rectangle (6.6,2.9) (2.6,2.45) -- (2.6,2.9)
(2.95,2.45) -- (2.95,2.9);
\node[font=\scriptsize] at (1.3,2.675) {committed $L_{a,m}(j)$};
\node[font=\scriptsize] at (2.775,2.675) {$j$};
\node[font=\scriptsize] at (4.775,2.675) {remaining working weights $\mathcal F$};
\node[draw, rounded corners=2pt, fill=blue!8, inner sep=1.7pt, font=\scriptsize]
(q) at (2.775,3.65) {shared code $q_j^{\star}$};
\draw[->] (q) -- (2.775,2.9);
\draw[->, dashed] (5.7,2.9) to[out=90, in=0] node[pos=.5, above
right=-2pt] {scores $q^{\star}_{j+1}$} (q.east);
\node[anchor=west] at (0,1.98) {\textbf{shared-state} (sh)};
\node[anchor=west] (inb) at (0,1.62) {$\Delta\mathbf W_{\mathcal F,1..4}$};
\node[draw, circle, inner sep=0.7pt, font=\scriptsize] (agg) at
(2.35,1.62) {avg};
\node[num] at (2.75,1.98) {1};
\draw[->] (inb) -- (agg);
\fill[gray!30] (2.95,1.41) rectangle (6.6,1.83); \draw
(2.95,1.41) rectangle (6.6,1.83);
\node[font=\scriptsize] at (4.775,1.62) {$\mathbf W^{\mathrm{sh}}_{\mathcal F}$ (shared)};
\draw[->] (agg) -- (2.95,1.62);
\node[anchor=west] at (0,1.15) {\textbf{\methodname{}} (MS)};
\foreach \m/\y in {1/0.80,2/0.39,3/-0.02,4/-0.43}{
\node[anchor=west] (inm\m) at (0,\y)
{$\Delta\mathbf W_{\mathcal F,\m}$};
\fill[blue!12] (2.95,\y-0.20) rectangle (6.6,\y+0.20); \draw
(2.95,\y-0.20) rectangle (6.6,\y+0.20);
\node[font=\scriptsize,yshift=-0.3pt, inner sep=0pt] at (4.775,\y)
{$\mathbf W^{(\m)}_{\mathcal F}$ (mode \m)};
\draw[->] (inm\m.east) -- (2.95,\y);}
\draw[<->] (6.8,1.62) -- node[right=-1pt, num] {2} (6.8,-0.43);
\end{tikzpicture}
\end{minipage}\hfill
\begin{minipage}[c]{0.56\textwidth}
\centering
\begin{tikzpicture}
\node[anchor=south west,inner sep=0] (plot) at (0,0)
{\includegraphics[width=\linewidth]{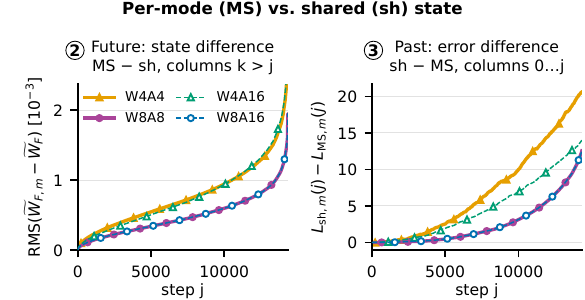}};
\begin{scope}[x={(plot.south east)},y={(plot.north west)}]
\fill[white] (0.034,0.130) rectangle (0.085,0.750);
\fill[white] (0.516,0.205) rectangle (0.569,0.675);
\node[rotate=90,font=\fontsize{8.0}{8.7}\selectfont] at (0.0595,0.441)
{$D_m(j)\,[10^{-3}]$};
\node[rotate=90,font=\fontsize{8.0}{8.7}\selectfont] at (0.5445,0.441)
{$L_{\mathrm{sh},m}(j)-L_{\mathrm{MS},m}(j)$};
\end{scope}
\end{tikzpicture}
\end{minipage}
\caption{One PTQ step with one shared working state against one working state per mode (Sec.~\ref{subsec:mode_consistent}). On the left, the GPTQ pass of Sec.~\ref{subsec:background_gptq} is at input index $j$ of one layer. Columns $0,\ldots,j-1$ are already quantized, column $j$ receives the joint code $q_j^{\star}$, and the working state of the remaining columns $\mathcal F$ scores the next code. \textcircled{\scriptsize 1} The baseline averages the four updates, whereas \methodname{} keeps them apart. \textcircled{\scriptsize 2} The shared working state thus drifts from that of each mode, plotted on the right as $D_m(j)$. \textcircled{\scriptsize 3} The drift changes the future quantized weights and raises the committed loss, plotted as $L_{\mathrm{sh},m}(j)-L_{\mathrm{MS},m}(j)$. Both follow Eq.~\eqref{eq:fig_diagnostics} for layer~31 \texttt{down\_proj} of Llama-3.1-8B-Instruct.}
\label{fig:compensation}
\end{figure*}

%% file: results/ladder_ppl_table.tex
\begin{tabular}{llrrrr}
\toprule
Flows & States & W4A4 & W4A16 & W8A8 & W8A16 \\
\midrule
\makebox[2.6em][l]{\makebox[0.55em][c]{$\bullet$}}shared & \makebox[2.6em][l]{\makebox[0.55em][c]{$\bullet$}}shared & 9.604 & 7.819 & 7.329 & 7.322 \\
\makebox[2.6em][l]{\makebox[0.55em][c]{$\bullet$}}shared & \makebox[2.6em][l]{\makebox[0.55em][c]{$\bullet$}\makebox[0.55em][c]{$\bullet$}\makebox[0.55em][c]{$\bullet$}\makebox[0.55em][c]{$\bullet$}}per mode & 9.093 & 7.573 & 7.261 & 7.256 \\
\makebox[2.6em][l]{\makebox[0.55em][c]{$\bullet$}\makebox[0.55em][c]{$\bullet$}\makebox[0.55em][c]{$\bullet$}\makebox[0.55em][c]{$\bullet$}}per mode & \makebox[2.6em][l]{\makebox[0.55em][c]{$\bullet$}}shared & 9.141 & 7.670 & 7.381 & 7.376 \\
\makebox[2.6em][l]{\makebox[0.55em][c]{$\bullet$}\makebox[0.55em][c]{$\bullet$}\makebox[0.55em][c]{$\bullet$}\makebox[0.55em][c]{$\bullet$}}per mode & \makebox[2.6em][l]{\makebox[0.55em][c]{$\bullet$}\makebox[0.55em][c]{$\bullet$}\makebox[0.55em][c]{$\bullet$}\makebox[0.55em][c]{$\bullet$}}per mode & \resMatched{8.791} & \resMatched{\resBest{7.528}} & \resMatched{7.256} & \resMatched{7.252} \\
\midrule
\multicolumn{2}{l}{Dedicated GPTAQ / GPTQ} & \resBest{8.496} & 7.598 & \resBest{7.224} & \resBest{7.219} \\
\bottomrule
\end{tabular}

%% file: sections/05_experiments.tex
\section{Flexibility and Accuracy}
\label{sec:experiments}
\label{sec:setup}

This section first shows what one checkpoint offers beyond its four fixed modes and then measures what nesting costs against quantizing one model per mode and against other nested methods.

\subsection{Experimental Setup}
\label{sec:quality_setup}

\paragraph{Models and format.} We quantize Llama-3.1-8B-Instruct~\cite{llama3} and Qwen3-8B~\cite{yang2025qwen3technicalreport}, the latter with thinking disabled. All \methodname{} modes share one scale $s_8$ per output channel, which maps the integer codes of W8 and W4 back to the range of the original weights and thereby sets the stride of their discrete quantization grids. As in MatGPTQ, we optimize it jointly for all modes by an MSE grid search, but weight the error of each mode by the diagonal of its own Hessian $\mathbf H_m$ instead of one shared Hessian. Per-channel scales follow GPTAQ and are typical for W4A4, whereas weight-only methods use group-wise scales. Activations are quantized per token at runtime and clipped to 0.9 of their range in the A4 modes as in GPTAQ, and the KV cache stays in 16 bits.

\paragraph{Calibration.} All methods calibrate on the same 256 sequences of 2048 tokens from FineWeb-Edu~\cite{penedo2024finewebedu}, so that they differ only in the method. All GPTAQ-based runs use GPTAQ's default $\alpha=0.25$ (Eq.~\eqref{eq:gptaq_mode_update}).

\paragraph{Compared methods.} Dedicated models quantize one model per mode and show what each mode reaches without nesting. We use GPTAQ~\cite{li2025gptaq}, adapted to Llama-3.1 and Qwen3, for W4A4 and W8A8 and GPTQ~\cite{frantar2023optq} from llm-compressor~\cite{llmcompressor2024} for W4A16 and W8A16.
Round-to-nearest (RTN) rounds every weight to its nearest grid point without calibration and serves as a lower reference, and BF16 shows the accuracy before quantization. Among nested methods, we compare against every method for LLMs with released code that calibrates without gradient training and serves at least two of our four modes from one stored integer code. These are MatGPTQ~\cite{kleinegger2026matgptqv2} and Any-Precision LLM~\cite{park2024anyprecisionllm}, which serve W4A16 and W8A16.

\paragraph{Evaluation.} We evaluate with the LM Evaluation Harness~\cite{gao2023lmeval} at batch size 1 on common-sense reasoning (CSR, the mean of ARC-Challenge, ARC-Easy~\cite{clark2018arc}, HellaSwag~\cite{zellers2019hellaswag}, PIQA~\cite{bisk2020piqa} and WinoGrande~\cite{sakaguchi2020winogrande}), GSM8K~\cite{cobbe2021gsm8k}, MMLU~\cite{hendrycks2020mmlu} and IFEval~\cite{zhou2023ifeval}, and report their mean as Avg.
Sections~\ref{sec:flexibility} and~\ref{sec:systems} additionally measure accuracy and speed in served vLLM runs on GSM8K, the harder MATH-500~\cite{hendrycks2021math,lightman2023verify} and five LongBench-E subtasks~\cite{bai2024longbench} (Qasper, HotpotQA, passage retrieval, SAMSum and TREC) with inputs of up to 4k words, 100 requests each, scored with their official metrics and averaged. All evaluations and timings use NVIDIA RTX 4090 GPUs.

\subsection{Flexibility of One Checkpoint}
\label{sec:flexibility}

One checkpoint can prefill a request in one mode and decode it in another, so it offers more choices than its four fixed modes. We call such a combination a precision switch and write it as X$\to$Y. The precision switch W4A4$\to$W4A16, for example, prefills with 4-bit activations and decodes with 16-bit activations, both with the 4-bit weights. Table~\ref{tab:switch} measures these choices for single requests, where every forward pass is either prefill or decode. For each task, it lists the change in accuracy and the speedup in output throughput over BF16 of every fixed mode and every precision switch, from the same runs of Llama-3.1-8B-Instruct with our vLLM integration (Sec.~\ref{sec:kernels}) on an RTX 4090. MATH-500 and GSM8K generate long solutions from short prompts, whereas long input processes long prompts, scored as the mean of five LongBench-E subtasks.

\input{figures/switch_table}
\begin{table*}[t]
\caption{Accuracy of the nested checkpoint against dedicated models in all four modes, with nested references and BF16 for context. Accuracy (\%) in the setting of Sec.~\ref{sec:quality_setup}. Bold marks the best quantized method.}
\label{tab:main}
\centering
\fontsize{8pt}{8.9pt}\selectfont \input{results/main_table}
\end{table*}

Decoding with 16-bit activations recovers accuracy without costing speed. The W4A4$\to$W4A16 switch runs at least as fast as fixed W4A4 and gains 13.8 points on MATH-500 and 6.1 on GSM8K over fixed W4A4, which closes the full gap to W4A16 on MATH-500 and about half of it on GSM8K.

Long input shows the limit of this switch. W4A4$\to$W4A16 recovers almost nothing there, and even decoding with the 8-bit weights after a W4A4 prefill still loses 3.3 points. The errors thus arise while processing the prompt and propagate to decoding. Processing it with the 8-bit weights avoids them, and fixed W8A8 is 0.2 points above BF16 at 1.67 times its speed.

Reading the input with the 8-bit weights and generating with the 4-bit weights combines both observations. It is 0.1 points above BF16 on long input at 1.96 times its speed and stays within 2.6 points of BF16 on all three tasks at 1.96 to 2.61 times its speed, which no fixed mode of the checkpoint reaches. Weight-only nested checkpoints such as MatGPTQ only offer the two WxA16 rows of Table~\ref{tab:switch}. Fully disaggregated quantization~\cite{panferov2026disaggregated} and FlexQuant~\cite{liu2025flexquant} also prefill and decode with different weights, but store both sets of weights.

\subsection{Comparison with Dedicated and Nested References}
\label{sec:results}

Table~\ref{tab:main} shows that the nested checkpoint loses little accuracy against dedicated models except at W4A4. In both W8 modes, \methodname{} scores between 0.24 points below and 0.22 points above BF16 on both models and at most 0.62 points below the dedicated models. At W4A16, \methodname{} is 0.46 points below dedicated GPTQ on Qwen and 0.39 points above it on Llama, with \resLlamaNestMSQWFourASixteenAvg{} against \resLlamaDedWFourASixteenAvg{}. IFEval reacts strongly to small changes of the weights. Its standard error is about 1.9 points on Llama, and dedicated W8A16 GPTQ spans 2.74 points over three calibration seeds. We therefore read the IFEval score of W4A16 RTN on Llama, 2.53 points above BF16, as an outlier. The gap concentrates at W4A4, where dedicated GPTAQ leads by 1.29 points on Llama and 1.01 on Qwen. Nesting prevents the W4 view from fitting its weight scale for 4 bits alone, as the dedicated W4A4 model does. The dedicated model also quantizes weights and activations with zero points, so part of the gap may come from this difference in format.

MatGPTQ and Any-Precision LLM provide only WxA16 modes. MatGPTQ, with finer group-wise scales, reaches about the accuracy of \methodname{} there, and Any-Precision LLM does not exceed it. Neither can vary the activation precision or use low-bit arithmetic from the same checkpoint. Whether these choices hold when many requests are processed in parallel depends on the kernels and the scheduler, which Sec.~\ref{sec:systems} examines.

%% file: figures/switch_table.tex
\begin{table}[t]
\caption{Switching precision between prefill and decoding of single requests. The BF16 row gives the reference score (\%). All other rows give the change in points ($\Delta$) and the speedup in output throughput relative to it. A precision switch X$\to$Y prefills in mode X and decodes in mode Y. Every speed comes from one timed run.}
\label{tab:switch}
\centering
\fontsize{7.5pt}{8.6pt}\selectfont \setlength{\tabcolsep}{2.4pt}\input{results/switch_table}
\end{table}

%% file: results/switch_table.tex
\begin{tabular}{@{}lrrrrrr@{}}
\toprule
 & \multicolumn{2}{c}{MATH-500} & \multicolumn{2}{c}{GSM8K} & \multicolumn{2}{c}{Long input} \\
\cmidrule(lr){2-3}\cmidrule(lr){4-5}\cmidrule(lr){6-7}
Mode & $\Delta$ & speed & $\Delta$ & speed & $\Delta$ & speed \\
\midrule
BF16 (\%) & 47.2 & 1.00 & 85.3 & 1.00 & 63.0 & 1.00 \\
\cmidrule(l){1-7}
W4A4 & $-$20.0 & 2.53 & $-$14.3 & 2.54 & $-$4.2 & 2.67 \\
W4A16 & $-$6.8 & 2.57 & $-$2.7 & 2.34 & $-$0.6 & 1.12 \\
W8A8 & $-$0.8 & 1.64 & +0.2 & 1.64 & +0.2 & 1.67 \\
W8A16 & +1.8 & 1.61 & $-$0.4 & 1.50 & +0.2 & 0.80 \\
\cmidrule(l){1-7}
W4A4$\to$W4A16 & $-$6.2 & 2.62 & $-$8.2 & 2.63 & $-$4.0 & 2.67 \\
W4A4$\to$W8A16 & $-$8.6 & 1.65 & $-$7.5 & 1.68 & $-$3.3 & 2.15 \\
W8A8$\to$W4A16 & $-$2.6 & 2.61 & $-$0.9 & 2.55 & +0.1 & 1.96 \\
W8A8$\to$W8A16 & $-$1.0 & 1.64 & +0.1 & 1.64 & +0.4 & 1.66 \\
\bottomrule
\end{tabular}

%% file: results/main_table.tex
\begin{tabular}{llrrrrr!{\modelsep}rrrrr}
\toprule
 &  & \multicolumn{5}{c}{Llama-3.1-8B-Instruct} & \multicolumn{5}{c}{Qwen3-8B} \\
\cmidrule(lr){3-7} \cmidrule(lr){8-12}
Mode & Method & CSR & GSM8K & MMLU & IFEval & Avg & CSR & GSM8K & MMLU & IFEval & Avg \\
\midrule
BF16 & BF16 & 73.70 & 85.97 & 69.41 & 77.19 & 76.57 & 71.45 & 87.41 & 75.00 & 83.30 & 79.29 \\
\midrule
W8A16 & RTN & \resLlamaRtnWEightASixteenCsr & \resLlamaRtnWEightASixteenGsm & \resLlamaRtnWEightASixteenMmlu & \resLlamaRtnWEightASixteenIfeval & \resLlamaRtnWEightASixteenAvg & \resQwenRtnWEightASixteenCsr & \resBest{\resQwenRtnWEightASixteenGsm} & \resQwenRtnWEightASixteenMmlu & \resBest{\resQwenRtnWEightASixteenIfeval} & \resBest{\resQwenRtnWEightASixteenAvg} \\
W8A16 & GPTQ & \resLlamaDedWEightASixteenCsr & \resLlamaDedWEightASixteenGsm & \resLlamaDedWEightASixteenMmlu & \resLlamaDedWEightASixteenIfeval & \resLlamaDedWEightASixteenAvg & \resQwenDedWEightASixteenCsr & \resQwenDedWEightASixteenGsm & \resQwenDedWEightASixteenMmlu & \resQwenDedWEightASixteenIfeval & \resQwenDedWEightASixteenAvg \\
W8A16 & MatGPTQ & \resLlamaMatGptqWEightASixteenCsr & \resLlamaMatGptqWEightASixteenGsm & \resBest{\resLlamaMatGptqWEightASixteenMmlu} & \resBest{\resLlamaMatGptqWEightASixteenIfeval} & \resLlamaMatGptqWEightASixteenAvg & \resQwenMatGptqWEightASixteenCsr & \resQwenMatGptqWEightASixteenGsm & \resBest{\resQwenMatGptqWEightASixteenMmlu} & \resQwenMatGptqWEightASixteenIfeval & \resQwenMatGptqWEightASixteenAvg \\
W8A16 & Any-Precision LLM & \resBest{\resLlamaAnyPrecWEightASixteenCsr} & \resBest{\resLlamaAnyPrecWEightASixteenGsm} & \resLlamaAnyPrecWEightASixteenMmlu & \resLlamaAnyPrecWEightASixteenIfeval & \resLlamaAnyPrecWEightASixteenAvg & \resBest{\resQwenAnyPrecWEightASixteenCsr} & \resQwenAnyPrecWEightASixteenGsm & \resQwenAnyPrecWEightASixteenMmlu & \resQwenAnyPrecWEightASixteenIfeval & \resQwenAnyPrecWEightASixteenAvg \\
W8A16 & \methodname{} & 73.73 & 85.60 & 69.11 & 78.50 & \resBest{76.74} & 71.59 & 87.57 & 74.87 & 84.01 & 79.51 \\
\midrule
W8A8 & RTN & \resLlamaRtnWEightAEightCsr & \resLlamaRtnWEightAEightGsm & \resLlamaRtnWEightAEightMmlu & \resLlamaRtnWEightAEightIfeval & \resLlamaRtnWEightAEightAvg & \resQwenRtnWEightAEightCsr & \resQwenRtnWEightAEightGsm & \resQwenRtnWEightAEightMmlu & \resBest{\resQwenRtnWEightAEightIfeval} & \resBest{\resQwenRtnWEightAEightAvg} \\
W8A8 & GPTAQ & \resBest{\resLlamaDedWEightAEightCsr} & \resBest{\resLlamaDedWEightAEightGsm} & \resBest{\resLlamaDedWEightAEightMmlu} & \resBest{\resLlamaDedWEightAEightIfeval} & \resBest{\resLlamaDedWEightAEightAvg} & \resQwenDedWEightAEightCsr & \resBest{\resQwenDedWEightAEightGsm} & \resBest{\resQwenDedWEightAEightMmlu} & \resQwenDedWEightAEightIfeval & \resQwenDedWEightAEightAvg \\
W8A8 & \methodname{} & 73.65 & 84.46 & 69.01 & 78.19 & 76.33 & \resBest{71.74} & 87.11 & 74.76 & 83.28 & 79.22 \\
\midrule
W4A16 & RTN & \resLlamaRtnWFourASixteenCsr & \resLlamaRtnWFourASixteenGsm & \resLlamaRtnWFourASixteenMmlu & \resBest{\resLlamaRtnWFourASixteenIfeval} & \resLlamaRtnWFourASixteenAvg & \resQwenRtnWFourASixteenCsr & \resQwenRtnWFourASixteenGsm & \resQwenRtnWFourASixteenMmlu & \resQwenRtnWFourASixteenIfeval & \resQwenRtnWFourASixteenAvg \\
W4A16 & GPTQ & \resLlamaDedWFourASixteenCsr & \resLlamaDedWFourASixteenGsm & \resLlamaDedWFourASixteenMmlu & \resLlamaDedWFourASixteenIfeval & \resLlamaDedWFourASixteenAvg & \resQwenDedWFourASixteenCsr & \resQwenDedWFourASixteenGsm & \resQwenDedWFourASixteenMmlu & \resBest{\resQwenDedWFourASixteenIfeval} & \resBest{\resQwenDedWFourASixteenAvg} \\
W4A16 & MatGPTQ & \resLlamaMatGptqWFourASixteenCsr & \resLlamaMatGptqWFourASixteenGsm & \resLlamaMatGptqWFourASixteenMmlu & \resLlamaMatGptqWFourASixteenIfeval & \resLlamaMatGptqWFourASixteenAvg & \resBest{\resQwenMatGptqWFourASixteenCsr} & \resBest{\resQwenMatGptqWFourASixteenGsm} & \resBest{\resQwenMatGptqWFourASixteenMmlu} & \resQwenMatGptqWFourASixteenIfeval & \resQwenMatGptqWFourASixteenAvg \\
W4A16 & Any-Precision LLM & \resLlamaAnyPrecWFourASixteenCsr & \resLlamaAnyPrecWFourASixteenGsm & \resLlamaAnyPrecWFourASixteenMmlu & \resLlamaAnyPrecWFourASixteenIfeval & \resLlamaAnyPrecWFourASixteenAvg & \resQwenAnyPrecWFourASixteenCsr & \resQwenAnyPrecWFourASixteenGsm & \resQwenAnyPrecWFourASixteenMmlu & \resQwenAnyPrecWFourASixteenIfeval & \resQwenAnyPrecWFourASixteenAvg \\
W4A16 & \methodname{} & \resBest{73.06} & \resBest{82.49} & \resBest{68.30} & 76.30 & \resBest{75.04} & 71.16 & 85.37 & 73.97 & 83.73 & 78.56 \\
\midrule
W4A4 & RTN & \resLlamaRtnWFourAFourCsr & \resLlamaRtnWFourAFourGsm & \resLlamaRtnWFourAFourMmlu & \resLlamaRtnWFourAFourIfeval & \resLlamaRtnWFourAFourAvg & \resQwenRtnWFourAFourCsr & \resQwenRtnWFourAFourGsm & \resQwenRtnWFourAFourMmlu & \resQwenRtnWFourAFourIfeval & \resQwenRtnWFourAFourAvg \\
W4A4 & GPTAQ & \resBest{\resLlamaDedWFourAFourCsr} & \resBest{\resLlamaDedWFourAFourGsm} & \resBest{\resLlamaDedWFourAFourMmlu} & \resBest{\resLlamaDedWFourAFourIfeval} & \resBest{\resLlamaDedWFourAFourAvg} & \resBest{\resQwenDedWFourAFourCsr} & \resBest{\resQwenDedWFourAFourGsm} & \resBest{\resQwenDedWFourAFourMmlu} & \resQwenDedWFourAFourIfeval & \resBest{\resQwenDedWFourAFourAvg} \\
W4A4 & \methodname{} & 69.10 & 71.19 & 61.63 & 72.59 & 68.63 & 66.39 & 78.32 & 68.43 & \resBest{82.21} & 73.84 \\
\bottomrule
\end{tabular}

%% file: sections/06_systems.tex
\section{Running the Four Modes from One Checkpoint}
\label{sec:systems}

A shared checkpoint pays off only if its modes run nearly as fast as the kernels that serve dedicated models of each precision, just as Sec.~\ref{sec:results} measures its accuracy against dedicated models. We show that all four modes run directly from the common INT8 weights with little overhead (Sec.~\ref{sec:kernels}) and that switching between them pays off end to end in vLLM~\cite{vllm}, for single requests and under continuous batching (Sec.~\ref{sec:serving_sb}), all on an RTX 4090.

\subsection{Shared Representation and Operator Efficiency}
\label{sec:kernels}

\input{figures/kernel_benchmark}

The kernels of the four modes have to share one weight representation in memory. We therefore build one kernel for each mode, and all four read the same stored parent code (Eq.~\eqref{eq:w4_w8}). The WxAx kernels additionally quantize the activations to 4 or 8 bits within timed kernel calls. For W4A4, this quantization also sums the 4-bit activations of each token in the same pass, and the matrix multiplication uses these sums in its epilogue to add the midpoint 7.5 of the W4 view to the integer result. The kernels build on Humming~\cite{humming2026} for W4A4, on the W8A8 kernel of OmniServe~\cite{lin2025qserve} and on Marlin~\cite{frantar2025marlin} for W4A16 and W8A16.

The layout of the stored code decides which kernel pays for reading it. As in BitNest~\cite{yang2026bitnest}, we store the coarse codes $C$ and the refinement codes $R$ in two separate planes, one tensor holding all $C$ and one holding all $R$, so the W4 kernels read only the $C$ plane. Each kernel expects its weights in the register order of its Tensor Core instruction and, for the A16 kernels, of Marlin's conversion to BF16. These orders differ, so the stored order can match only one kernel. The W8 kernels have to merge $C$ and $R$ into $q=(C\ll4)\,|\,R$ by a shift and a bitwise OR in any case, so the matching order goes to a W4 kernel. We give it to W4A4, the W4 mode used at large $M$, because its compute-bound INT4 multiplication leaves the least time to hide extra instructions, whereas W4A16 runs memory-bound at small $M$ and hides them behind the weight loads. Its kernel loads $C$ into registers unchanged. The other three kernels rearrange $C$ on the CUDA cores, and the W8 kernels also merge it with $R$ there. Meanwhile, asynchronous copies load the next tiles and the Tensor Cores multiply the current ones, so loading, rearranging and multiplying run in parallel.

Figure~\ref{fig:kernels} plots the speedup of each kernel over BF16 against the number of tokens $M$ of a matrix multiplication, for the four projection shapes of Llama-3.1-8B. The A16 kernels are faster at $M=1$ and the WxAx kernels at large $M$, with a crossover between 8 and 64 tokens that depends on the matrix shape. The kernels that vLLM runs for each precision, shown as hollow markers with their own layouts, show the same crossover. Where each mode is used, our kernels keep most of the reference kernels' throughput. The A16 kernels reach at least 94\,\% of Marlin up to 16 tokens, and from 512 tokens on, W8A8 reaches at least 85\,\% of CUTLASS and W4A4 at least 88\,\% of Humming. From 256 tokens on, even Marlin is at most 1.16 times as fast as BF16, whereas our W4A4 and W8A8 kernels are 2.8 to 6.2 and 1.3 to 2.3 times as fast as BF16. Our A16 kernels fall behind Marlin there, because the stored layout is ordered for W4A4 and they have to rearrange the weights. W4A16 reaches 82 to 87\,\% of Marlin and W8A16 60 to 73\,\%. Weight packing happens before timing.

\subsection{Serving with Continuous Batching}
\label{sec:serving_sb}
\label{sec:serving_mb}

\input{figures/fullmodel_serving}
\input{figures/serving_load}

Sec.~\ref{sec:flexibility} switches precision within single requests. Under continuous batching, the load decides which of the modes pays off, and the one resident checkpoint serves each of them without loading another model. Under load, vLLM batches the running requests continuously~\cite{vllm}. Each iteration is one forward pass of the model over a budget of at most 2048 tokens. It takes one decode token from every running request and fills the rest of the budget with prompt tokens of waiting requests, which it splits into chunks if a prompt does not fit. Prompt chunks of new requests and decode tokens of running ones therefore often share an iteration. We vary the maximum number of running requests, $K$.

A precision switch therefore chooses its mode per iteration, with a deliberately simple rule that needs neither the context length nor a cost model. An iteration that contains prompt tokens runs the prefill mode, so that the prompt is always read in the precision on which the accuracy on long input depends (Sec.~\ref{sec:flexibility}). Decode tokens in such an iteration run in the prefill mode as well. A decode-only iteration holds one token per running request. It runs the decode weights with 16-bit activations if it holds at most 32 tokens and with low-bit activations otherwise. With few tokens, loading the weights limits its speed, and 16-bit activations cost almost nothing. With many tokens, the arithmetic limits it, and low-bit activations pay off. We locate this boundary by timing decoding iterations of the full model in every precision mode. Up to 32 tokens, an A16 iteration costs at most 5\,\% more than a WxAx iteration at every measured context. Beyond, the WxAx modes are faster by an amount that depends on the context. The 32 tokens are a conservative threshold, not a general crossover.

How much of the decoding runs with 16-bit activations therefore depends on the load. We call the fraction of decode tokens that a switch computes with 16-bit activations its A16 share. Our rule lowers it below 100\,\% whenever decode tokens share an iteration with prompt tokens or more than 32 requests decode at once. Figures~\ref{fig:fullmodel_serving} and~\ref{fig:serving_load} measure the speed in controlled runs with fixed prompt and output lengths, first for single requests and then under load. Figure~\ref{fig:recovery} and Table~\ref{tab:load_switch} measure the accuracy in natural runs on the tasks of Sec.~\ref{sec:flexibility}.

\paragraph{Single requests.} For single requests, a switch is at most 0.6\,\% slower than either of its two fixed modes, from prefill-heavy to decode-heavy requests. Figure~\ref{fig:fullmodel_serving} measures the switches of Sec.~\ref{sec:flexibility} end to end over the phase mix of single requests, against the ratio $\rho=G/P$ of generated to prompt tokens, from prefill-heavy requests at $\rho=1/128$ to decode-heavy ones at $\rho=1$. At W4, the W4A4$\to$W4A16 switch runs each phase in its faster mode and is at least as fast as both fixed modes in all 28 measured combinations of $P$ and $\rho$. At W8, fixed W8A8 is as fast as fixed W8A16 for decode-heavy requests, so the W8A8$\to$W8A16 switch decodes with 16-bit activations at the speed of fixed W8A8. At the decode-heavy end, the dedicated W4A16 and W8A16 models are up to 4.6\,\% faster than the best mode or switch of the shared checkpoint with the same weights, because Marlin reads their weights in its native layout, while our layout favors W4A4 (Sec.~\ref{sec:kernels}).

\paragraph{Speed under load.} Figure~\ref{fig:serving_load} serves batches of $K=16$, 32 and 128 requests with prompts of $P=128$ tokens over the ratio $\rho=G/P$ from $1/32$ to 8. Its top panel shows the A16 share, and its bottom panel the throughput relative to fixed W4A4. At low load, almost all decode tokens run with A16, and the share falls for short outputs and when more than 32 requests decode at once. The switch nevertheless keeps 0.96 to 1.05 times the throughput of fixed W4A4 at every load, so the decode tokens it computes with 16-bit activations come at almost no cost in speed. Fixed W4A16 with our kernels instead reaches at most 63\,\% of the throughput of fixed W4A4 at $K=128$.

\paragraph{Accuracy under load.} A lower A16 share costs no speed, but it lowers the accuracy that the switch recovers. Figure~\ref{fig:recovery} plots, for each task and load, how much of the W4A4 accuracy gap to W4A16 the W4 switch recovers, against its A16 share. On MATH-500, the recovery grows almost in proportion to the share. On GSM8K it reaches about half of it, and on long input it stays near zero, as for single requests. In the natural runs of Table~\ref{tab:load_switch}, new requests keep arriving, and their prompts share iterations with the decoding of the others. Table~\ref{tab:load_switch} gives the resulting A16 shares.

\paragraph{Reading with 8 bits under load.} Where decoding with 16-bit activations recovers little, reading the prompt with the 8-bit weights keeps the accuracy. Table~\ref{tab:load_switch} gives accuracy and speed of every fixed mode and precision switch under load, at one load per task. On long input, W4A4$\to$W4A16 stays 4.1 points below BF16, because decoding with 16-bit activations cannot repair the errors made while W4A4 reads the prompt. The precision switch W8A8$\to$W4A16 instead reads the prompt in W8A8 and decodes in W4A16. It is 0.5 points above BF16 on long input at 1.72 times its speed and stays within 2.1 points of BF16 on all three tasks at 1.72 to 2.36 times its speed. At the same loads, it reaches 1.02 to 1.64 times the output throughput of the dedicated GPTQ W4A16 model and scores 1.1 to 3.0 points higher, but none of these accuracy differences is statistically significant at the 5\,\% level (paired bootstrap over requests). The dedicated model cannot prefill in W8A8, so serving this switch with dedicated models takes a W8A8 model next to it, whereas \methodname{} reads both from one INT8 checkpoint.

\input{figures/load_switch_table}
\input{figures/recovery}

\paragraph{Scope.} The threshold of 32 tokens comes from the cost of a decoding iteration of Llama-3.1-8B-Instruct with our kernels on the RTX 4090 and has to be measured again for other GPUs, kernels and models. We keep the rule simple because it only has to show that one checkpoint can follow the phase and the load. Finer policies that choose the precision per layer, per token or over the course of decoding~\cite{kwon2025dpllm,wang2026mobiquant,chen2025pmpd} could run on the same checkpoint.

\FloatBarrier

%% file: figures/kernel_benchmark.tex
\begin{figure*}[t]
\centering
\includegraphics[width=\textwidth]{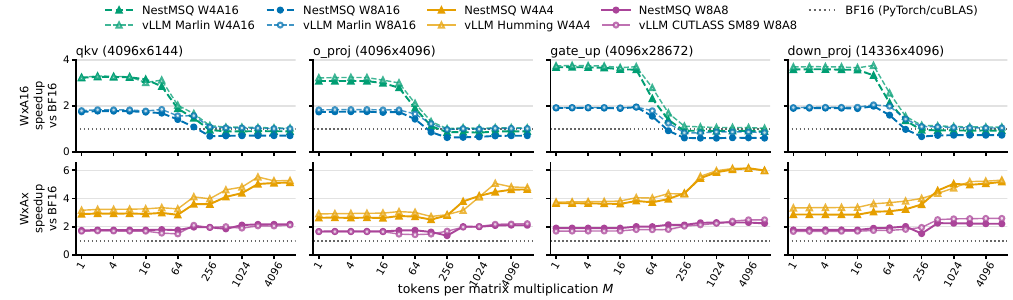}
\caption{Operator execution from one resident nested checkpoint. Projection speedup over BF16 versus the number of tokens per matrix multiplication, $M$, for the four Llama-3.1-8B projection shapes. Top, W4A16/W8A16. Bottom, W4A4/W8A8. Filled markers execute from the shared representation, hollow markers are reference kernels with their own layouts, vLLM Marlin (A16), vLLM CUTLASS (W8A8) and vLLM Humming (W4A4). The measured crossover between WxA16 and WxAx depends on projection shape and weight precision rather than following a single universal $M$ threshold. Medians of three sessions, RTX~4090.}
\label{fig:kernels}
\end{figure*}

%% file: figures/fullmodel_serving.tex
\begin{figure}[t]
\centering
\includegraphics[width=\columnwidth]{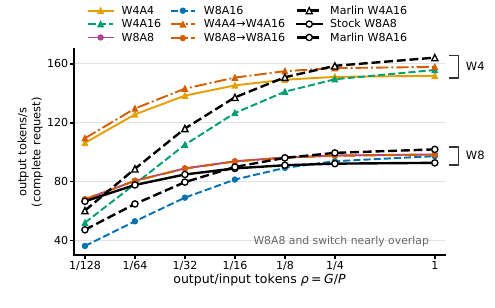}
\caption{Throughput of single requests from prefill-heavy to decode-heavy workloads. Output tokens per second of complete requests of Llama-3.1-8B-Instruct on an RTX 4090 over the ratio $\rho=G/P$ of generated to prompt tokens, for the fixed modes, the two switches and the dedicated models. Medians of three sessions.}
\label{fig:fullmodel_serving}
\end{figure}

%% file: figures/serving_load.tex
\begin{figure}[t]
\centering
\includegraphics[width=\columnwidth]{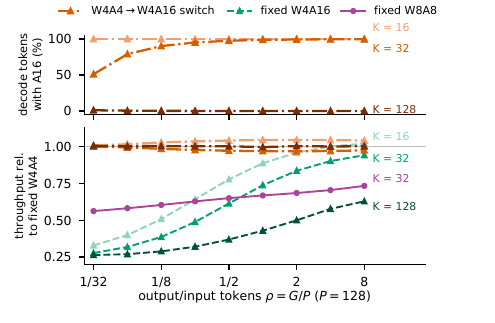}
\caption{The precision switch W4A4$\to$W4A16 as the load grows. Batches of $K$ requests with $P=128$ prompt tokens and equal $G$ start together, with $K$ shown by shade, on an RTX 4090. Top, the A16 share of the switch. Bottom, throughput relative to fixed W4A4 at the same $K$ and $\rho$, with fixed W4A16 and, at $K=32$, fixed W8A8. Means of three sessions, fixed W4A16 one session.}
\label{fig:serving_load}
\end{figure}

%% file: figures/load_switch_table.tex
\begin{table}[tb]
\caption{The same precision switches with $K$ concurrent requests per task. The BF16 row gives the reference score (\%) at the same load. All other rows give the change in points ($\Delta$) and the speedup in output throughput relative to it. The GPTQ rows are dedicated models, quantized separately with GPTQ, one per precision. A switch X$\to$Y prefills in mode X and decodes in mode Y. With the per-iteration rule of Sec.~\ref{sec:serving_sb}, the switches decode 98 to 99\,\% of their tokens with 16-bit activations on MATH-500, 74 to 77\,\% on GSM8K and 38 to 45\,\% on long input. Every cell comes from one run.}
\label{tab:load_switch}
\centering
\fontsize{7.5pt}{8.6pt}\selectfont \setlength{\tabcolsep}{1.8pt}\resizebox{\columnwidth}{!}{\input{results/load_switch_table}}
\end{table}

%% file: results/load_switch_table.tex
\begin{tabular}{@{}lrrrrrr@{}}
\toprule
 & \multicolumn{2}{c}{MATH-500, $K=8$} & \multicolumn{2}{c}{GSM8K, $K=32$} & \multicolumn{2}{c}{Long input, $K=6$} \\
\cmidrule(lr){2-3}\cmidrule(lr){4-5}\cmidrule(lr){6-7}
Mode & $\Delta$ & speed & $\Delta$ & speed & $\Delta$ & speed \\
\midrule
BF16 (\%) & 45.8 & 1.00 & 85.7 & 1.00 & 62.7 & 1.00 \\
\cmidrule(l){1-7}
W4A4 & $-$18.4 & 2.38 & $-$15.1 & 2.24 & $-$4.0 & 2.61 \\
W4A16 & $-$5.6 & 2.15 & $-$2.8 & 1.02 & $-$0.4 & 0.89 \\
W8A8 & +4.0 & 1.61 & $-$0.2 & 1.56 & +0.4 & 1.71 \\
W8A16 & +1.8 & 1.45 & $-$0.3 & 0.77 & +0.2 & 0.67 \\
GPTQ W4A16 & $-$4.8 & 2.31 & $-$3.2 & 1.23 & $-$0.9 & 1.05 \\
GPTQ W8A16 & $-$0.2 & 1.57 & $-$0.3 & 1.05 & +0.6 & 0.93 \\
\cmidrule(l){1-7}
W4A4$\to$W4A16 & $-$7.6 & 2.43 & $-$10.1 & 2.22 & $-$4.1 & 2.64 \\
W4A4$\to$W8A16 & $-$7.8 & 1.64 & $-$10.0 & 1.93 & $-$4.8 & 2.55 \\
W8A8$\to$W4A16 & $-$1.8 & 2.36 & $-$2.0 & 1.72 & +0.5 & 1.72 \\
W8A8$\to$W8A16 & +1.2 & 1.61 & $-$0.9 & 1.53 & +0.4 & 1.71 \\
\bottomrule
\end{tabular}

%% file: figures/recovery.tex
\begin{figure}[tb]
\centering
\includegraphics[width=\columnwidth]{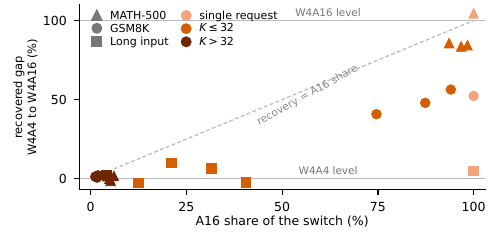}
\caption{Accuracy recovered by the precision switch W4A4$\to$W4A16 against its A16 share. Recovery is $(\text{switch}-\text{W4A4})/(\text{W4A16}-\text{W4A4})$, with all three scores from natural runs at the same task and $K$.}
\label{fig:recovery}
\end{figure}

%% file: sections/07_discussion.tex
\section{Discussion and Limitations}
\label{sec:discussion}

\paragraph{Scope.} We demonstrate \methodname{} on two 8B models for accuracy and on Llama-3.1-8B-Instruct with one 24\,GB RTX 4090 for speed, where decoding is limited by weight loading and the choice of activation precision matters most. With more memory, more requests run at once, so the A16 share has to be measured again for each GPU and model.

\paragraph{Number format and scales.} We start with integer formats, because their bits nest cleanly. The 4-bit code is the upper half of the 8-bit code, and both share one scale per output channel that is applied once after integer accumulation. Attention and the KV cache stay in 16 bits. Newer GPUs move their low-bit arithmetic to floating point. Hopper lists INT8 and FP8 but no INT4 Tensor Core format~\cite{nvidia2026h100}, and Blackwell adds 4-bit floating point~\cite{nvidia2026gb200}. Nesting floating-point formats is harder, because their low-order bits act on the value multiplicatively~\cite{kleinegger2026matgptqv2}, and nested floating-point formats~\cite{lee2026nestedfp} show one way. Group-wise scales also improve accuracy in W4A4 quantization~\cite{ashkboos2024quarot} but add work to the main loop of the integer matrix multiplication, which requires applying a scale for every group~\cite{lin2025qserve}. The per-mode flows and working states carry over to both extensions, because they do not depend on the number format.

\paragraph{Policy and calibration cost.} The checkpoint provides the choices, and Secs.~\ref{sec:flexibility} and~\ref{sec:systems} measure what each is worth. Which choice to make for a request depends on its task and on what the deployment optimizes. A policy that chooses per request, or per token as for WxA16 models~\cite{wang2026mobiquant}, can build on these measurements. Four calibration flows take \resCalibTimeMsGptaq{} for Llama-3.1-8B-Instruct on one RTX~4090, against \resCalibTimeSingleFlow{} for one shared flow. This cost is paid once per model, and the per-mode working states need one copy of the current layer per mode during calibration. Neither adds to the stored checkpoint or to the inference latency.

%% file: sections/08_conclusion.tex
\section{Conclusion}
\label{sec:conclusion}

We presented \methodname{}, which shares the deployed nested code across the four modes W4A4, W8A8, W4A16 and W8A16 but keeps a calibration flow and a working state per mode during PTQ. The resulting INT8 checkpoint loses at most 1.3 points of average accuracy against dedicated models in every mode and runs all four modes directly from its stored code. Their speed differences persist from single requests to continuous batching, so one checkpoint can prefill and decode a request in different modes. Reading the input with the 8-bit weights and generating with the 4-bit weights stays within 2.6 points of BF16 for single requests and within 2.1 points under load, at 1.72 to 2.61 times the output throughput of BF16. Choosing among the modes for a given workload remains a separate policy problem.